\documentclass[letterpaper, 10 pt, conference]{ieeeconf}  

\IEEEoverridecommandlockouts                              

\usepackage{amsfonts}
\usepackage{graphicx}
\usepackage{textcomp}
\usepackage{xcolor}
\usepackage{url}
\usepackage{float}

\usepackage{comment}
\usepackage{algpseudocode}
\usepackage{xcolor}
\usepackage{times}
\usepackage{epsfig}
\usepackage{amsmath}
\usepackage{amssymb}
\usepackage{siunitx}
\usepackage{color}
\usepackage{booktabs}
\usepackage{algorithm}
\usepackage{cite}
\usepackage[font=small,labelfont=bf]{caption}
\usepackage{subcaption}
\usepackage{caption, subcaption}
\usepackage{array}

\usepackage{multirow}
\usepackage{graphicx}
\let\labelindent\relax
\usepackage{enumitem}
\setlist[itemize]{align=parleft,left=0pt}

\newcommand{\figref}[1]{Fig.~\ref{#1}}

\usepackage[switch]{lineno}

\def\BibTeX{{\rm B\kern-.05em{\sc i\kern-.025em b}\kern-.08em
    T\kern-.1667em\lower.7ex\hbox{E}\kern-.125emX}}
    
\makeatletter
\let\NAT@parse\undefined
\makeatother

\usepackage[pagebackref=true,breaklinks=true,colorlinks,bookmarks=false,citecolor=green]{hyperref}
\begin{document}

\title{FusionPainting: Multimodal Fusion with Adaptive Attention for \\3D Object Detection
}

\author{Shaoqing Xu$^{1, 2}$, Dingfu Zhou$^{2, 3}$, Jin Fang$^{2, 3}$, Junbo Yin$^{2, 4}$, Zhou Bin$^{1\ *}$ and Liangjun Zhang$^{2, 3}$  

\thanks{$^{1}$ Beihang University, Beijing 100083, China. $^{2}$ Robotics and Autonomous Driving Laboratory, Baidu Research. $^{3}$ National Engineering Laboratory of Deep Learning Technology and Application, China. $^{4}$ Beijing Institute of Technology, Beijing, China.}
\thanks{
        {\tt\small \{xushaoqing, zhoudingfu, fangjin, liangjun zhang\}@baidu.com} 
        }%
\thanks{
        \tt\small \{xsq0226,binzhou\}@buaa.edu.cn
        }%
\thanks{
        *Corresponding author.
        }%
}

\maketitle
\begin{abstract}

Accurate detection of obstacles in 3D is an essential task for autonomous driving and intelligent transportation. In this work, we propose a general multimodal fusion framework \textit{FusionPainting} to fuse the 2D RGB image and 3D point clouds at a semantic level for boosting the 3D object detection task. Especially, the \textit{FusionPainting} framework consists of three main modules: a multi-modal semantic segmentation module, an adaptive attention-based semantic fusion module, and a 3D object detector. First, semantic information is obtained for 2D image and 3D Lidar point clouds based on 2D and 3D segmentation approaches. Then the segmentation results from different sensors are adaptively fused based on the proposed attention-based semantic fusion module. Finally, the point clouds painted with the fused semantic label are sent to the 3D detector for obtaining the 3D objection results. The effectiveness of the proposed framework has been verified on the large-scale nuScenes detection benchmark by comparing with three different baselines. The experimental results show that the fusion strategy can significantly improve the detection performance compared to the methods using only point clouds, and the methods using point clouds only painted with 2D segmentation information. Furthermore, the proposed approach outperforms other state-of-the-art methods on the nuScenes testing benchmark. Code will be available at {\url{ https://github.com/Shaoqing26/FusionPainting/}}. 

\end{abstract}

\section{Introduction}\label{sec:intro}
3D object detection, as a fundamental task in computer vision and robotics, has been extensively studied with the development of autonomous driving and intelligent transportation. Currently, LiDAR sensor is widely used for perception tasks because it can provide accurate range measurements of the surrounding environment, especially for the obstacles such as vehicles, pedestrians and cyclists, etc. With the development of the deep learning techniques on point-cloud based representation, many LiDAR-based 3D object detection approaches have been developed. Generally, these approaches can be categorized into point-based \cite{shi2019pointrcnn} and voxel-based \cite{zhou2018voxelnet, lang2019pointpillars} methods. LiDAR sensors have the superiority of providing distance information of the obstacles, even though the detailed geometry is often lost due to its sparse scanning and furthermore texture and color information can be not captured. Therefore, False Positive (FP) detection and wrong categories classification often happen for LiDAR-based object detection solutions.

\begin{figure}[ht!]
	\centering
	\includegraphics[width=0.45\textwidth]{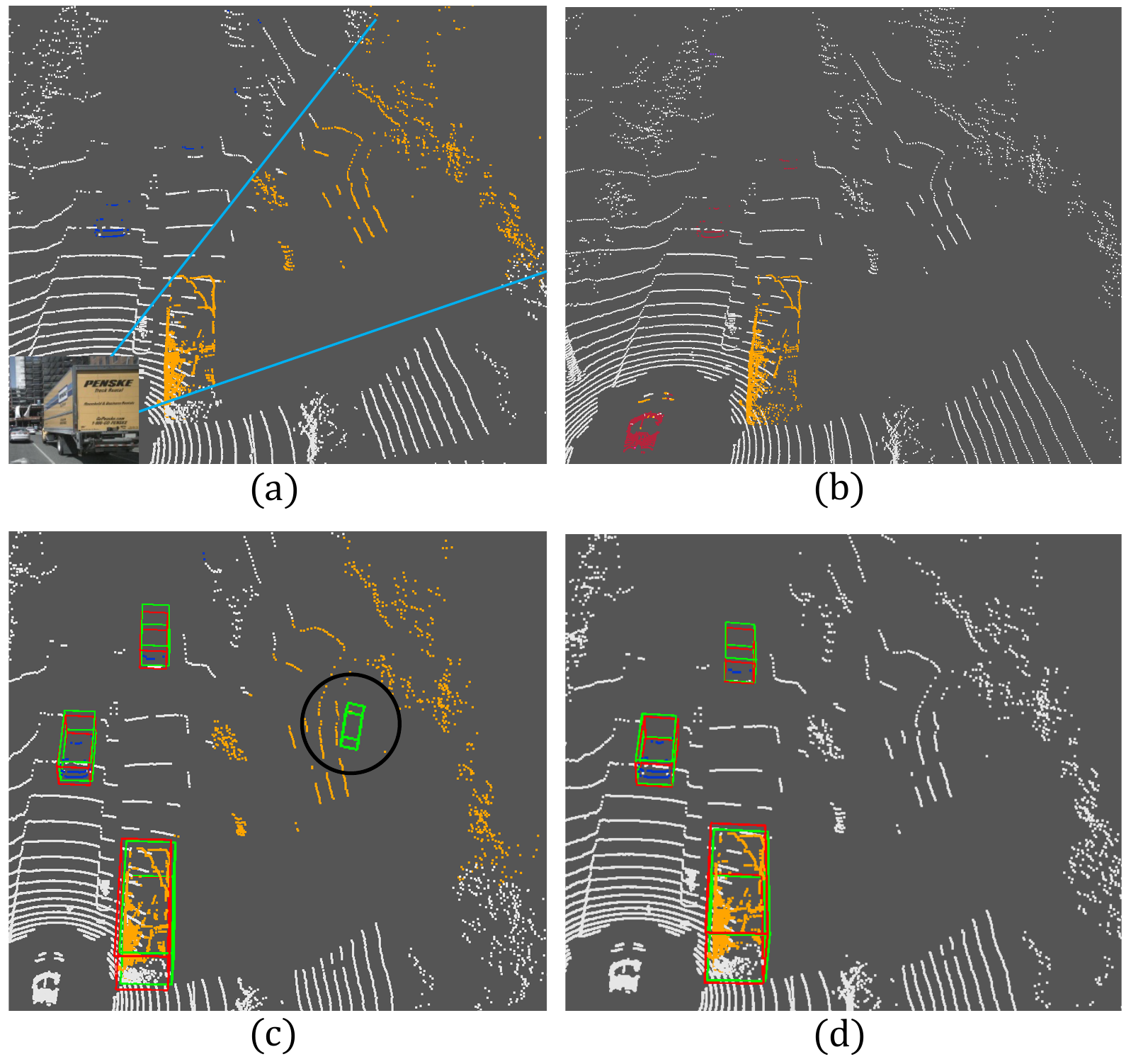}
	\centering
	\caption{(a) Painted pointcloud with the 2D segmentation results. The frustum within the blue line shows misclassified area due to the boundary-blurring effect; (b) Painted point cloud by 3D segmentation results (misclassified points are demonstrated in red color); (c) and (d) give the object detection results with False Positive(FP) based on 2D painted point cloud (a) and the proposed 2D/3D fusion framework, respectively.} 
	\label{Fig:2DSeg_in_3d}
\end{figure}

On the contrary, the camera sensors can provide detailed texture and color information with high resolution, though the depth has been lost during the perspective projection based imaging procedure. The combination of the two different types of sensors of LiDAR and camera is a promising way for boosting the performance of autonomous driving perception. In literature, multi-modal based object detection approaches can be divided into as early fusion \cite{dou2019seg, vora2020pointpainting}, deep fusion \cite{zhang2020maff, xu2018pointfusion, tian2020adaptive} and late fusion approaches \cite{pang2020clocs}. Early fusion approaches aim at creating a new type of data by combining the raw data directly before sending them into the detection framework. Usually, these kinds of methods require pixel-level correspondence between each type sensor data. Different from the early fusion methods, late fusion approaches execute the detection for each type of data separately first and then fuse the detection results in the bounding box level. Different from the above two methods, deep fusion-based methods usually extract the features with different types of deep neural networks first and then fuse them at the features level. As a simple yet effective sequential fusion method, PointPainting \cite{vora2020pointpainting} has achieved superior detection results on different benchmarks. This approach employees the 2D image semantic segmentation results from an off-the-shelf neural network first and then adds them into a point-cloud-based 3D object detection based on the 2D-3D projection. The superiority of PointPainting suggests that  2D image segmentation approach can be used for providing the semantic results and it can be incorporated into any 3D object detectors even point-based or voxel-based approaches.

However, the boundary-blurring effect often happens in image-based semantic segmentation methods due to the relatively low resolution of the deep feature map. This effect becomes much more severe after re-projecting them into the 3D point cloud. An example of the reprojected 2D result into 3D is shown in sub-fig. \ref{Fig:2DSeg_in_3d}-(a). Taking the big truck at the bottom of the image as an example, we can find that there is a large frustum area of the background (e.g, points in orange color) that has been miss-classified as foreground due to the inaccurate segmentation results in the 2D image. In addition, the correspondence of 3D points to 2D image pixels is not exactly a one-to-one projection due to the digital quantization problem, and many-to-one projection issues. An interesting phenomenon is that the segmentation from the 3D point cloud (e.g., sub-fig. \ref{Fig:2DSeg_in_3d}-(b)) performs much better on the boundary of obstacles. However, compared to the 2D image, the category classification from the 3D point cloud often gives worse results(e.g.,point in blue color) due to the detailed texture information from the RGB images.

The painted points \cite{vora2020pointpainting}, with semantic information has been proved to be very effective for the object detection task even with some semantic errors. An intuition idea is that the detection performance can be further improved if 2D and 3D segmentation results can be fused together. Based on this idea, we propose a general multi-modal fusion framework ``FusionPainting'' to fuse different types of sensors at the semantic level to further boost the 3D object detection performance. First, any off-the-shelf semantic segmentors are employed directly for obtaining the semantic information. 
To further improve the performance, an attention module is proposed to fuse different kinds of semantic information in voxel-level adaptively based on the learned context features in a self-attention style. Finally, the points painted by the fused semantic labels are sent to 3D detectors for obtaining the final detection results. 
The main contributions of this work can be summarized as following:
\begin{enumerate}
   \item A general multi-modal fusion framework ``FusionPainting'' has been firstly proposed to fuse the different types of information in the semantic level to improve the 3D object detection performance.
   \item To further improve the performance, an attention module is proposed to fuse different kinds of semantic information in voxel-level by learning the context features. 
  \item The experimental results on the large-scale autonomous driving benchmark nuScenes show the superiority of the proposed fusion framework and achieve SOTA results compared to other methods. 
\end{enumerate}


\section{Related Work}\label{sec:related_work}

\subsection{3D Object Detection in Point Cloud} 
The existing point cloud-based 3D object detection methods can be categorised as projection-based~\cite{ku2018joint,luo2018fast,yang2018pixor,liang2020rangercnn,hu2020you}, voxel-based~\cite{zhou2018voxelnet,yan2018second,kuang2020voxel, yin2020lidar,zhou2019iou} and point-based~\cite{zhou2020joint,lang2019pointpillars,qi2018frustum}. RangeRCNN \cite{liang2020rangercnn} is a typical 3D object detection framework which is based on the range image, and generates anchors on the BEV (bird’s-eye-view) map. Different from hand-crafted features design in previous works, VoxelNet\cite{zhou2018voxelnet} proposes a VFE (voxel feature encoding) layer, which can learn a unified feature representation for each 3D voxel. Based on VoxelNet, CenterPoint\cite{yin2021center} presents an anchor-free method which use a center-based framework based on CenterNet \cite{duan2019centernet}, and achieves state-of-the-art performance. SECOND\cite{yan2018second} leverages the sparse convolution \cite{graham20183d} operation to alleviate the burden of 3D convolution. PointPillars\cite{lang2019pointpillars} extracts the features from vertical columns (Pillars) with PointNet\cite{qi2017pointnet} and encodes the features as a pseudo image, then 2D object detection pipeline can be applied. PointRCNN \cite{shi2019pointrcnn} is a pioneer work that directly generates 3D proposals from a raw point cloud. Furthermore, PV-RCNN \cite{shi2020pv} integrates the multi-scale 3D voxel CNN and PointNet-based network to learn more discriminative features.


\begin{figure*}[ht!]
	\centering
	\includegraphics[width=0.85\textwidth]{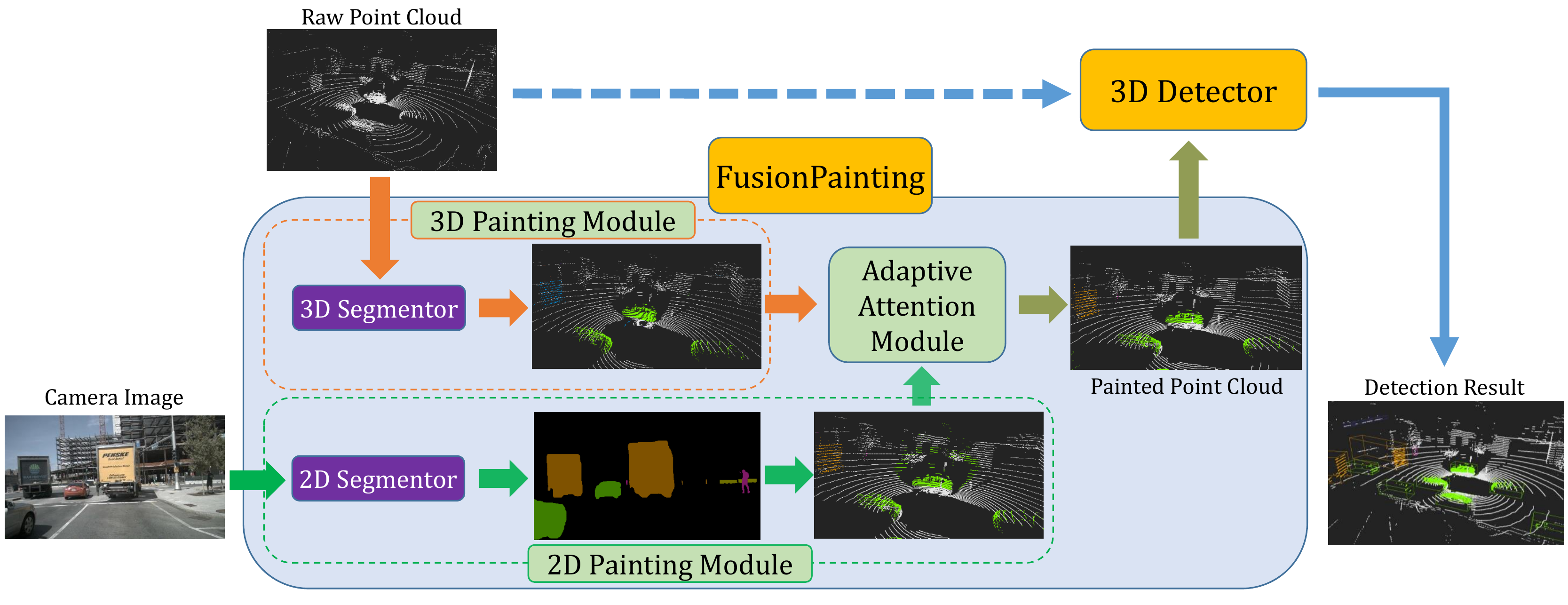}
	\centering
	\caption{Overview of the proposed ``FusionPainting'' framework. We first the process the input point cloud and 2D image with 2D and 3D segmentors to obtain the semantic estimation. Then, the proposed adaptive attention module is adopted to fuse the two types of the sensor at the semantic level. Finally, the painted point cloud is sent to modern 3D detectors to produce the detection results.}
	\label{Fig:PointPainting}
\end{figure*}

\subsection{Multi-modal Fusion}

Various methods exploit combining multiple types of sensory data to boost the detection performance. LiDAR and camera are the most used sensors. MV3D \cite{chen2017multi} proposes a multi-view representation including BEV, camera view and front view image. The framework generates proposals from the BEV and deeply fuses the features from other views. AVOD \cite{ku2018joint} takes LiDAR point clouds
and RGB images as input, to generate features shared by RPN (region proposal network) and the refined network. F-PointNet \cite{qi2018frustum} uses the image to generate proposals and refines the final bounding box in cropped frustums with PointNet. Different from the above methods, PointPainting \cite{vora2020pointpainting} uses an easy but effective strategy: retrieving the class score vector for each point by projecting them into image semantic segmentation results, and feeding the concatenated attribution to 3D object detection framework. Multi-task fusion\cite{he2017mask} \cite{liang2019multi} is also an effective technology, e.g., \cite{zhou2020joint} jointing the semantic segmentation with 3D object detection tasks to further improve the performance. Furthermore, in \cite{yang2018hdnet} and \cite{fang2021mapfusion}, the HD Maps are also taken as an strong prior information for movable object detection in autonomous driving scenario.






\section{Proposed Approach} \label{sec:method}
We describe the architecture of our ``FusionPainting'' framework in \figref{Fig:PointPainting}. We propose to leverage both the 2D images and 3D point clouds information for obtaining accurate 3D locations of obstacles in the autonomous driving scenarios. Here, the fusion process is achieved by adaptively integrating 2D and 3D information at the semantic level. As shown in \figref{Fig:PointPainting}, the proposed framework consists of three models as a multi-modal semantic segmentation module, an adaptive attention module, and a 3D detector module. First, any off-the-shelf 2D and 3D segmentors can be employed to obtain the semantic segmentation from the RGB image and LiDAR point clouds respectively. Then a simple but effective attention strategy is proposed to sufficiently benefit the merits and suppress the drawbacks of the 2D and 3D semantic results. Finally, painted 3D point clouds with enhanced semantic labels are sent to modern 3D object detectors for achieving the final detection results.    



\subsection{Multi-modal Semantic Segmentation}

\textbf{2D Painting Module:} 2D images contain rich texture and color information, which can provide complementary clues for point clouds and thus benefit the 3D detection. To acquire useful information from 2D images, we utilize a semantic segmentation network to produce pixel-wise semantic labels. In particular, the proposed framework is agnostic to specific segmentation model, and any state-of-the-art approaches can be employed here, such as Deeplabv3 \cite{chen2018encoder}, PSPNet \cite{zhao2017pyramid}, HTC \cite{chen2019hybrid}, FCN \cite{shelhamer2017fully}, MobelNetV3 \cite{Howard_2019_ICCV} etc.. The segmentation network takes multi-view images as input and outputs pixel-wise classification labels for the foreground instances and the background. Here, we denote the 2D segmentation results as $ S \in \mathbb{R}^{w\times h \times m}$,  where $w$, $h$ are the width and length of the input image, $m$ is the number of classes. Following the sequential fusion strategy in ~\cite{vora2020pointpainting}, we reproject the pixel-wise 2D segmentation mask into the corresponding 3D point with the provided camera projection matrices. Assuming the intrinsic matrix is $\mathbf{K} \in \mathbb{R}^{4\times 4}$ and the extrinsic matrix is $\mathbf{M}\in \mathbb{R}^{3\times 4}$ and the original 3D points is $\mathbf{P}\in \mathbb{R}^{N\times 3}$, the projection of 3D point cloud from LiDAR coordinate to camera coordinate can be described as
\begin{equation}
	 {\mathbf{P}}^{'} = \text{Proj}(\mathbf{K}, \mathbf{M}, \mathbf{P}),
\end{equation} 
where $\mathbf{P}^{'}$ is the projected point in camera coordinate and the ``Proj'' represents the project process. With this projection, then the semantic labels in the 2D image can be assigned to the 3D point cloud. By presenting the pixel-wise semantic score in a one-hot encoding manner, then we can obtain the 2D painted points as $\mathbf{P}_{2D}\in \mathbb{R}^{N\times m}$. 


\textbf{3D Painting Module:} besides 2D semantic segmentation, we also consider obtaining the semantic information from LiDAR point clouds alternatively. Specifically, the 3D segmentation network Cylinder3D\cite{cong2021input} is employed directly to obtain the point-wise segmentation mask. Cylinder3D is a novelty 3D segmentation approach that uses cylindrical and asymmetrical 3D convolution networks for achieving SOTA performance. In addition, any SOTA 3D segmentators such as Polarnet \cite{zhang2020polarnet}, AF2S3Net \cite{cheng20212}, RPVNet \cite{xu2021rpvnet} can be used for the proposed framework. Furthermore, any extra point-wise semantic annotations are not required, we can freely use the 3D object bounding boxes annotations for generating the segmentation labels. For foreground instances, the points inside a 3D bounding box are directly assigned with the corresponding class label. And the points outside all the 3D bounding boxes are taken as the background. From this point of view, the proposed framework can work on any 3D detection datasets directly without any extra point-wise semantic annotations. After obtaining the trained network, the 3D painted points $\mathbf{P}_{3D}\in \mathbb{R}^{N\times m}$, can be obtained directly from the segmentation results.

\begin{figure*}[ht!]
	\centering
	\includegraphics[width=0.80\textwidth]{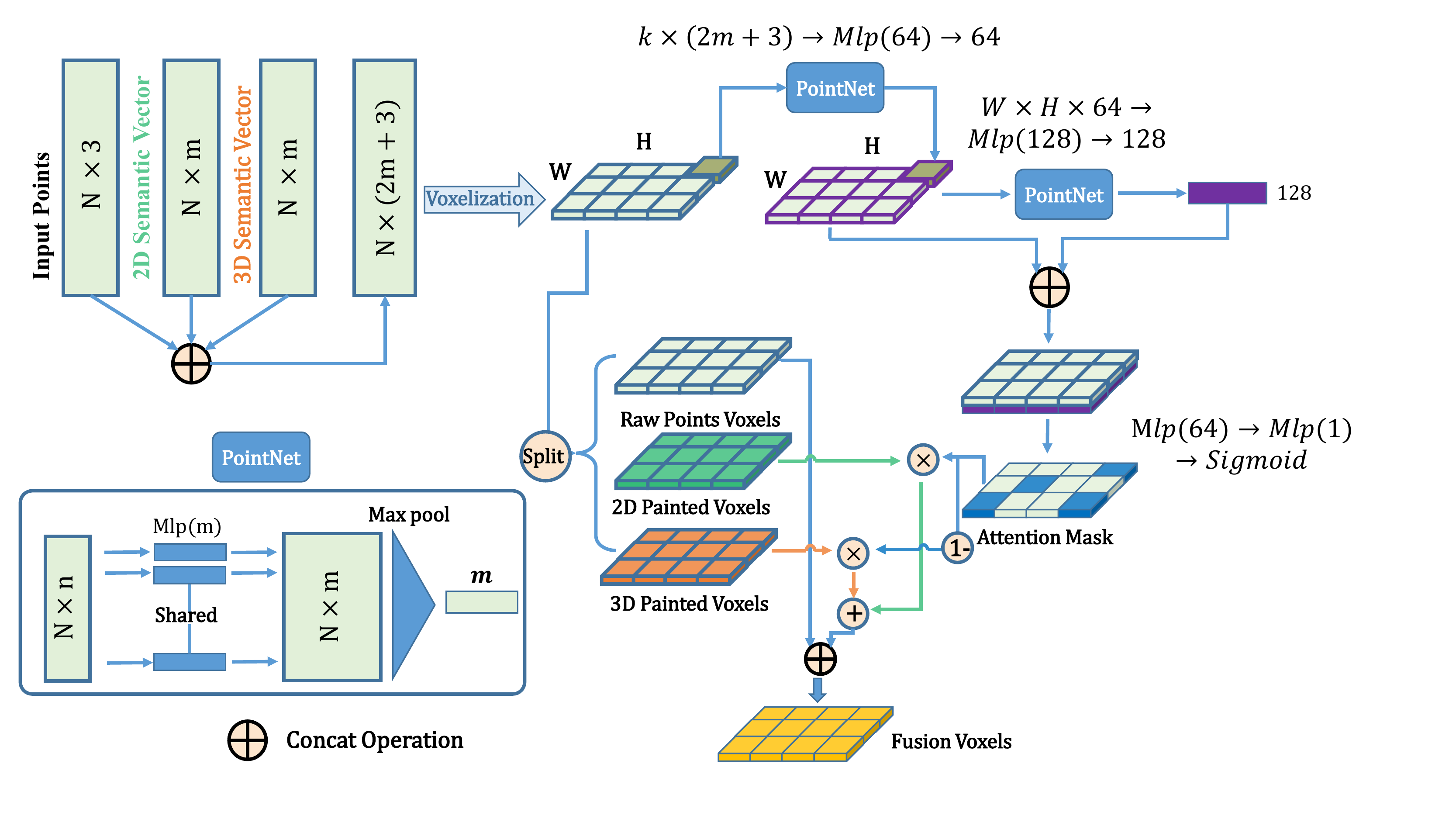}
	\centering
	\caption{The architecture of the proposed Adaptive Attention module. The input points and 2D/3D semantic predictions are first utilized to learn an attention mask. Then, the raw voxel is painted with 2D/3D semantic labels adaptive by using the learned attention mask.}
	\label{Fig:attention_module}
\end{figure*}
\subsection{Adaptive Attention Module}
Though the 2D segmentation achieves impressive performance, the boundary-blurring effect is also inevitable due to the limited resolution of the feature map. Therefore, the painted point cloud from the 2D segmentation mask usually has misclassified regions around the boundary of the objects. For example, the frustum region is illustrated in the sub-fig. ~\ref{Fig:2DSeg_in_3d} (a) behind the big truck. On the contrary, the point cloud-based semantic segmentation methods often can produce a clear and accurate object boundary e.g., sub-fig.~\ref{Fig:2DSeg_in_3d}(b). In order to combine the merits of both modules and suppress their disadvantages, we propose an adaptive attention module that adaptively fuses the semantic labels with an attention mechanism. By doing this, the refined semantic labels can further improve the following 3D detection results.


The detailed architecture of the proposed adaptive attention module is illustrated in Fig.~\ref{Fig:attention_module}. 
By defining the input point clouds as $\{ \mathbf{P}_i, i=1,2,3...N\}$, each point $\mathbf{P}_i$ contains $(x, y, z)$ coordinates and other features if they have, such as intensity and other attributes. For simplicity, only the coordinates $(x, y, z)$ is considered in the following context. Our goal is to find an optimal strategy to fuse semantic labels from the 2D segmentation $\mathbf{P}_{2D}$ and 3D segmentation $\mathbf{P}_{3D}$. Here, we proposed to utilize the learned attention mask to fuse the types of labels. Specifically, we first combine the points coordinates, attributes, and 2D/3D labels with a concatenation operation and obtain a fused point cloud with the size of $N\times (2m+3)$. For saving memory consumption, the following fusion is executed at the voxel level rather than the point level. Specifically, each point cloud has been divided into evenly distributed voxels and each voxel contains a fixed number of points. Here, we represent the voxels as $\{ V_i, i=1,2,3...E\}$, where $E$ is the total number of the voxels and each voxel $V_{i} \in \mathbb{R} ^{M \times(2m+3)}$ containing a fixed number of $M$ points with $2m+3$ features. Here, the sampling technique is adopted to keep the same number of points in each voxel as k. Then, local and global feature aggregation strategies are applied to estimate an attention weight for each voxel for determining the importance of the 2D and 3D semantic labels.


For the local feature, a PointNet~\cite{qi2017pointnet}-like module is employed here to extract the voxel-wise information inside each non-empty voxel. For $i$-th voxel, its local feature can be represented as 
\begin{equation}
    V_i = f(p_1, p_2,  \cdots , p_N) =
    \max_{i=1,...,N} \{\text{Mlp}_{l}(p_i)\} \in \mathbb{R}^{C_1}
\end{equation}
where $\text{Mlp}_{l}(\cdot)$ and ${max}$ are the local muti-layer perception (Mlp) networks and max-pooling operation. Specifically, $\text{Mlp}_{l}(\cdot)$ consists of a linear layer, a batch normalization layer, an activation layer and outputs local features with $C_1$ channels. To achieve global feature information, we aggregate information based on the $E$ voxels. In particular, we first use a $\text{Mlp}_{g}(\cdot)$ to map each voxel features from $C_1$ dimension to $C_2$ dimension. Then, another PointNet-like module is applied on all the voxels as 
\begin{equation}
     V_{global} = f(V_1, V_2, \cdots ,V_E) = \max_{i=1,...,N} \{\text{Mlp}_{g}(V_i)\} \in \mathbb{R}^{C_2}.
\end{equation}
Finally, the global feature $V_{global}$ is expaned to the same size of voxel number and then concatenate them to each local feature $V_i$ to obtain final fused local and global feature $V_{gl}\in \mathbb{R}^{N \times{(C_1 + C_2)}}$.

Then, the fused feature $V_{gl}$ is adopted to estimate an attention score for each voxel. This is achieved by applying another Mlp module $\text{Mlp}_{att}(\cdot)$ on $V_{gl}$ followed by a Sigmod activation function $\sigma(\cdot)$. Afterwards, we multiply the confidence score by corresponding one-hot semantic vectors for each point in a voxel, which is denoted as 
\begin{equation}
	    \mathbf{P}_{a.S}  = \mathbf{P}_{2D} \otimes \sigma{(h_{att}(V_{gl})}) \label{eq:2d_f},
\end{equation} 
\begin{equation}
	    \mathbf{P}_{a.T} = \mathbf{P}_{3D} \otimes (1-\sigma{(h_{att}(V_{gl}}))),
     \label{eq:3d_f}
\end{equation} 
where $\mathbf{P}_{2D}$ and $\mathbf{P}_{3D}$ are the 2D and 3D semantic segmentation labels and $\mathbf{P}_{a.S}$ and $\mathbf{P}_{a.T}$ are the corresponding enhanced features. Finally, we fuse $\{\mathbf{P}_{a.T}, \mathbf{P}_{a.S}\} \in \mathbb{R}^{N \times M \times m}$ with raw point $P$ in each voxel obtaining the enhance voxels $\widehat{V}_m\in \mathbb{R}^{N\times M \times (3 + 2m)}$, where $m, M$ denote semantic classes number and points number in voxel. Finally, $\hat{V_m}$ contains the adaptively fused information from 2D and 3D semantic labels.

\begin{table*}[ht!]
\renewcommand\arraystretch{1.3}
\centering
 \renewcommand\arraystretch{1.1}
\resizebox{0.95\textwidth}{!}
{
	\begin{tabular}{l | c | c | c c c c c c c c c c}
	\hline
    \multicolumn{1}{c|}{\multirow{2}{*}{Methods}} &
    \multicolumn{1}{c|}{\multirow{2}{*}{\textbf{NDS}(\%)}} &
    \multicolumn{1}{c|}{\multirow{2}{*}{\textbf{mAP}(\%)}} & 
    \multicolumn{10}{c}{\textbf{AP} (\%)} \\
    
    \multicolumn{1}{c|}{} & \multicolumn{1}{c|}{} & \multicolumn{1}{l|}{} & \multicolumn{1}{l}{Car} & \multicolumn{1}{l}{Pedestrian} & \multicolumn{1}{l}{Bus} & \multicolumn{1}{l}{Barrier} & \multicolumn{1}{l}{T.C.} &
    \multicolumn{1}{l}{Truck} & \multicolumn{1}{l}{Trailer} & \multicolumn{1}{l}{Moto.} & \multicolumn{1}{l}{Cons.} &
    \multicolumn{1}{l}{Bicycle} \\ \hline

	SECOND \cite{yan2018second}  &61.96 &50.85 &81.61  &77.37 & 68.53 &57.75  &56.86  &51.93 &38.19 & 40.14 &17.95 & 18.16 \\
	SECOND$^{\ast}$ & 67.41  & 62.15  & 83.98  & 82.86 &71.13 &63.29 &74.66 &59.97 &41.53 &66.85 &22.90  &54.35  \\
	Improvement $\uparrow$ & \textcolor{red}{\textbf{+5.45}} & \textcolor{red}{\textbf{+11.30}} & \textcolor{red}{+2.37} & \textcolor{red}{+5.49} & \textcolor{red}{+2.60} & \textcolor{red}{+5.54} & \textcolor{red}{+17.80} & \textcolor{red}{+8.04} & \textcolor{red}{+3.34} & \textcolor{red}{+26.71} & \textcolor{red}{+5.05}& \textcolor{red}{+36.19} \\
    \hline
	
	PointPillars \cite{lang2019pointpillars}  & 57.50 &43.46 &80.67 &70.80 &62.01 &49.23 &44.00 &49.35 &34.86 &26.74 &11.64 &5.27 \\
	PointPillars$^{\ast}$ &66.04 & 60.77&83.51 &82.90&69.96 &58.46&74.50&56.90&39.25&66.40&21.68&54.10   \\
	Improvement $\uparrow$ & \textcolor{red} {\textbf{+8.54}} & \textcolor{red}{\textbf{+17.31}} & 
	\textcolor{red}{+2.84} & \textcolor{red}{+12.10} & \textcolor{red}{+7.95 } & \textcolor{red}{ +9.23} & \textcolor{red}{+30.50} & \textcolor{red}{+7.55} & \textcolor{red}{+4.39} & \textcolor{red}{+39.66}  & \textcolor{red}{+10.04} & \textcolor{red}{+48.83} \\
    \hline
	
	CenterPoint \cite{yin2021center}  & 64.82 &56.53 &84.73 &83.42 &66.95 &64.56 &64.76 &54.52 &36.13 &56.81 &15.81 &37.57\\
	CenterPoint$^{\ast}$ &70.68 &66.53 &87.04 &88.44 & 70.66 &67.26 &79.57 &62.98 & 45.09 & 74.65&25.36&64.41\\
	Improvement $\uparrow$ & \textcolor{red}{\textbf{+5.86}} & \textcolor{red}{\textbf{+10.00}} & \textcolor{red}{+2.31} & \textcolor{red}{+5.02} & \textcolor{red}{+3.71} & \textcolor{red}{+2.70} & \textcolor{red}{+14.81} & \textcolor{red}{+8.46} & \textcolor{red}{+8.96} & \textcolor{red}{+17.84} &  \textcolor{red}{+9.55}&\textcolor{red}{+26.84} \\
    \hline
	\end{tabular}
}
\caption{\normalfont Evaluation results on nuScenes validation dataset. ``NDS'' and ``mAP'' mean nuScenes detection score and mean Average Precision. ``T.C.'', ``Moto.'' and ``Cons.'' are short for ``traffic cone'', ``motorcycle'', and ``construction vehicle'' respectively. `` * '' denotes the improved baseline by adding the proposed ``FusionPainting''. To hight the effectiveness of our methods, the improvements of each method are demonstrated in red.}
\label{tab:eval_on_nuscenes_val}
\end{table*}

\subsection{3D Object Detection Module}

After obtaining the painted point cloud, any off-the-shelf 3D object detectors can be directly employed for predicting 3D object detections. The 3D detector receives the painted voxels produced by the adaptive attention module, and achieve better results for all the categories. The detailed analysis of the performance can be found in the following experimental results part.

\begin{table*}[ht!]
    \centering
    \renewcommand\arraystretch{1.2}
\resizebox{0.95\textwidth}{!}{

\begin{tabular}{ l| c| c| c| c c c c c c c c c c}
    \hline
    \multicolumn{1}{c|}{\multirow{2}{*}{Methods}}  & 
    \multicolumn{1}{c|}{\multirow{2}{*}{Modality}} & \multicolumn{1}{c|}{\multirow{2}{*}{\textbf{NDS}(\%)}} & 
    \multicolumn{1}{c|}{\multirow{2}{*}{\textbf{mAP}(\%)}} & 
    \multicolumn{10}{c}{\textbf{AP} (\%)} \\ 
    \multicolumn{1}{c|}{} & \multicolumn{1}{c|}{} & \multicolumn{1}{c|}{} & \multicolumn{1}{c|}{} & \multicolumn{1}{l}{Car} & \multicolumn{1}{l}{Truck} & \multicolumn{1}{l}{Bus} & \multicolumn{1}{l}{Trailer} & \multicolumn{1}{l}{Cons} & \multicolumn{1}{l}{Ped} & \multicolumn{1}{l}{Moto} & \multicolumn{1}{l}{Bicycle} & \multicolumn{1}{l}{T.C} & \multicolumn{1}{l}{Barrier} \\ \hline
    PointPillars \cite{vora2020pointpainting} & L & 55.0 & 40.1 & 76.0 & 31.0 & 32.1 & 36.6 & 11.3 & 64.0 & 34.2 & 14.0 & 45.6 & 56.4 \\
    3DSSD \cite{yang20203dssd} & L  & 56.4 & 46.2 & 81.2  & 47.2  & 61.4  &  30.5 & 12.6 &  70.2 & 36.0  &  8.6 &  31.1 &  47.9 \\
    
    CBGS \cite{zhu2019class} & L & 63.3 & 52.8 & 81.1 & 48.5 & 54.9 & 42.9 & 10.5 & 80.1 & 51.5 & 22.3 & 70.9 & 65.7 \\
    HotSpotNet \cite{chen2020object} & L & 66.0 & 59.3 & 83.1  & 50.9 & 56.4  & 53.3  &  {23.0} &  81.3 & {63.5}  &  36.6 &  73.0 &  \textbf{71.6} \\
    CVCNET\cite{chen2020every} & L & 66.6 & 58.2 & 82.6 & 49.5 & 59.4 & 51.1 & 16.2 & {83.0} & 61.8 & {38.8} & 69.7 & 69.7 \\
    CenterPoint \cite{yin2021center} & L & {67.3} & {60.3} & {85.2} & {53.5} & {63.6} & {56.0} & 20.0 & 54.6 & 59.5 & 30.7 & ne{78.4} & 71.1 \\
    PointPainting \cite{vora2020pointpainting} & L \& C & 58.1 & 46.4 & 77.9 & 35.8 & 36.2 & 37.3 & 15.8 & 73.3 & 41.5 & 24.1 & 62.4 & 60.2 \\
    3DCVF \cite{DBLP:conf/eccv/YooKKC20} & L \& C & 62.3 & 52.7 & 83.0 & 45.0 & 48.8 & 49.6 & 15.9 & 74.2 & 51.2 & 30.4 & 62.9 & 65.9 \\ \hline
    {FusionPainting(Ours)} & L \& C & \textbf{70.4} & \textbf{66.3} & \textbf{86.3} & \textbf{58.5} & \textbf{66.8} & \textbf{59.4} & \textbf{27.7} & \textbf{87.5} & \textbf{71.2} & \textbf{51.7} & \textbf{84.2} & {70.2}  \\  \hline
    \end{tabular}
}
\caption{Comparison with other SOTA methods on the nuScenes 3D object detection benchmark. ``L'' and ``C'' in the modality column represent lidar and camera respectively. For each column, we use ``bold'' to illustrate the best results of other methods. Here, we only list all the methods results with publications.}
\label{tab:test_tabel}
\end{table*}

\section{Experimental Results} \label{sec:experiments}
We evaluate the effectiveness of the proposed ``FusionPainting'' on the large-scale autonomous driving 3D object detection dataset. First, the details of experiments are described and then the evaluation results on three baseline detectors are given.


\subsection{Detectors, Dataset and Implementation Details} \label{subsec:dataset}

\textbf{Baselines:} in addition, to verify its universality,we  have implemented the proposed module on three different State-of-the-Art 3D object detectors as following,
\begin{itemize}
\item \textit{SECOND} \cite{yan2018second}, which is the first to employ the sparse convolution on the voxel-based 3D object detection task and the detection efficiency has been highly improved compared to previous work such as VoxelNet. 
\item \textit{PointPillars} \cite{lang2019pointpillars}, which divides the point cloud into pillars and ``Pillar Feature Net'' is applied to each pillar for extracting the point feature. Then 2d convolution network has been adopted on the bird-eye-view feature map for object detection. PointPillars is a trade-off between efficiency and performance. 
\item \textit{CenterPoint} \cite{yin2021center}, is the first anchor-free based 3D object detector which is very suitable for small object detection. 
\end{itemize}


\textbf{Dataset:} nuScenes 3D object detection benchmark~\cite{nuscenes2019} has been employed for evaluation here, which is a large-scale dataset with a total of 1,000 scenes. For a fair comparison, the dataset has been officially divided into train, val, and testing, which includes 700 scenes (28,130 samples), 150 scenes (6019 samples), 150 scenes (6008 samples) respectively. For each video, only the key frames (every 0.5s) are annotated with 360-degree view. With a 32 lines LiDAR used by nuScenes, each frame contains about {300,000} points. For object detection task, 10 kinds of obstacles are considered including  ``car'', ``truck'', ``bicycle'' and ``pedestrian'' et al. Besides the point clouds, the corresponding RGB images are also provided for each keyframe. For each keyframe, there are 6 images that cover 360 fields-of-view.




\textbf{Evaluation Metrics:} For 3D object detection, \cite{nuscenes2019} proposes mean Average Precision (mAP) and nuScenes detection score (NDS) as the main metrics. Different from the original mAP defined in \cite{everingham2010pascal}, nuScenes consider the BEV center distance with thresholds of \{0.5, 1, 2, 4\} meters, instead of the IoUs of bounding boxes. NDS is a weighted sum of mAP and other metric scores, such as average translation error (ATE) and average scale error (ASE).

 
\textbf{Implementation Details:} HTCNet \cite{chen2019hybrid} and Cylinder3D \cite{zhou2020cylinder3d} are employed here as the 2D Segmentor and 3D Segmentor, respectively, due to their outstanding semantic segmentation ability. The 2D segmentor is pretrained on nuImages\footnote{\url{https://www.nuscenes.org/nuimages}} dataset, and the 3D segmentor is pretrained on nuScenes detection dataset while the point cloud semantic information can be generated by extracting the points inside each bounding box of the obstacles.
For Adaptive Attention Module, $m=11$, $C_1=64,C_2=128$, respectively, 
For each baseline, all the experiments share the same setting, the voxel size for SECOND, PointPillar and CenterPoint are $0.2m \times 0.2m \times 8 m$, $0.1m \times 0.1m \times0.2m$ and $0.075m \times 0.075m \times 0.075m$, respectively.
We use AdamW \cite{ loshchilov2019decoupled} with max learning rate 0.001 as the optimizer. Follow \cite{nuscenes2019}, 10 previous LiDAR sweeps are stacked into the keyframe to make the point clouds more dense.

\subsection{Evaluation Results} \label{subsec:evaluation}
We have evaluated the proposed framework on nuScenes benchmark for both validation and test splits.

\textbf{Evaluation on Baseline Methods:} First of all, we have integrated the proposed fusion module into three different baselines methods and experimentally we have achieved consistently improvements on both $\text{mAP}$ and $\text{NDS}$ compared to all baselines. Detailed results are given in Tab. \ref{tab:eval_on_nuscenes_val}. From this table, we can obviously find that the proposed module gives more than 10 points improvements on $\text{mAP}$ and  5 points improvements on $\text{NDS}$ for all the three baselines. The improvements have reached 17.31 and 8.45 points for PointPillars method  on $\text{mAP}$ and $\text{NDS}$ respectively. In addition, we find that the ``Traffic Cone'', ``Moto'' and ``Bicycle'' have received the surprising improvements compared to other classes. For ``Bicycle'' category specifically, the AP has improved about \textbf{36.19\%}, \textbf{48.83\%} and \textbf{26.84\%} based on Second, PointPillar and Centerpoint respectively. Interestingly, all these categories are small objects which are hard to be detected because of a few LiDAR points on them. By adding the prior semantic information, the category classification becomes relatively easier. Especially, the experiments here we share the same setting, so the baseline figures may be subtle differences with official.


\textbf{Comparison with other SOTA methods:} To compare the proposed framework with other SOTA methods, we submit our best results (proposed module on the CenterPoint \cite{yin2021center}) to the nuScenes evaluation sever\footnote{\url{https://www.nuscenes.org/object-detection/}}. The detailed results are listed in Tab. \ref{tab:test_tabel}. To be clear, only these methods with publications are compared here due to space limitation. From this table, we can find that the ``NDS'' and ``mAP'' have been improved by 3.1 points and 6.0 points respectively compared with the baseline method CenterPoint \cite{yin2021center}. More importantly, our algorithm outperforms previous multi-modal method 3DCVF \cite{DBLP:conf/eccv/YooKKC20} by a large margin, \textit{i.e.}, improving 8.1 and 13.6 points in terms of NDS and mAP metrics.


\begin{figure*}[ht!]
	\centering
	\includegraphics[width=0.975\textwidth]{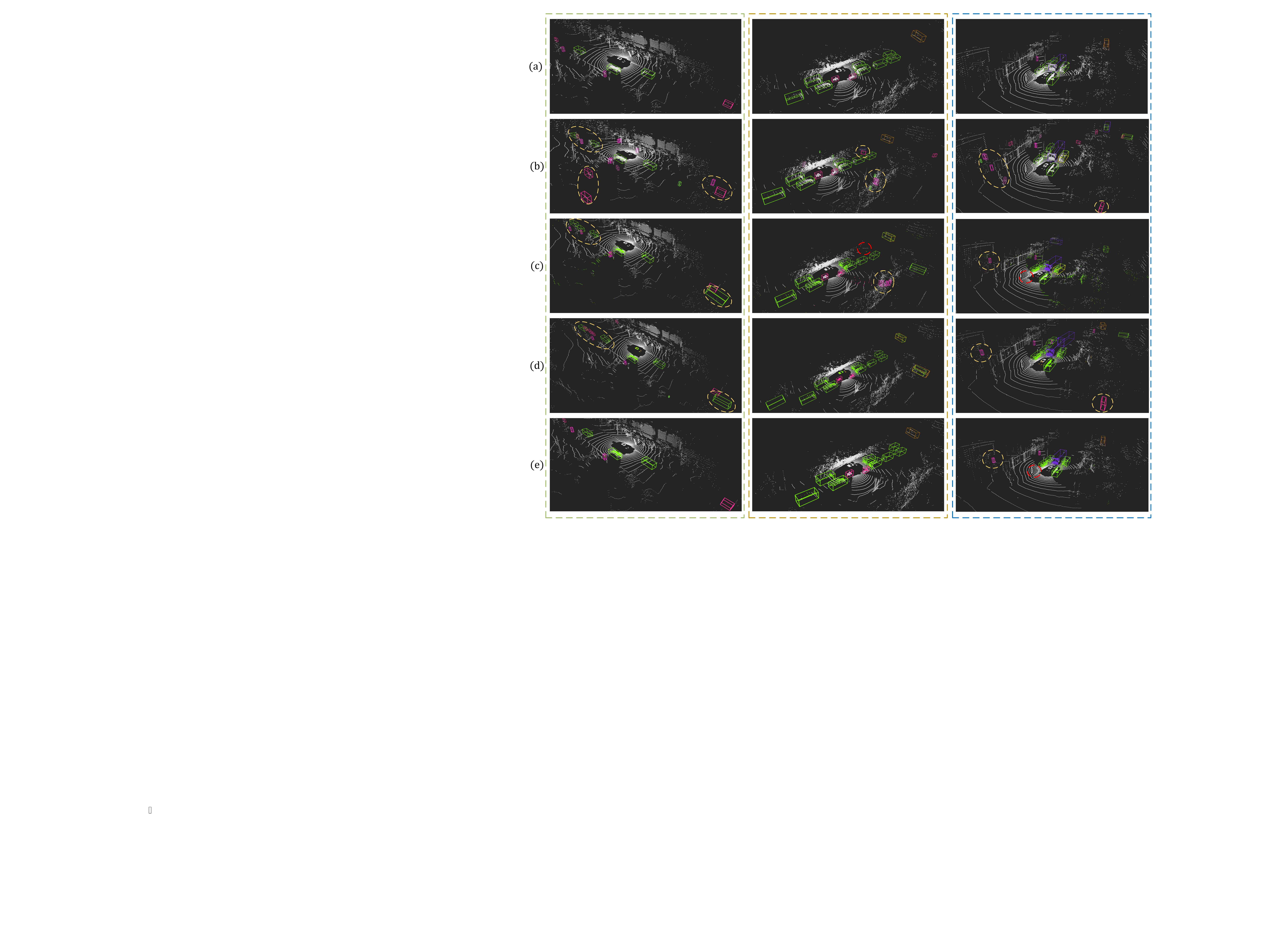}
	\centering
	\caption{Visualization of detection results, where (a) is the ground-truth, (b) is the detection results based on raw point cloud, (c), (d) and (e) are the detection results based on the projected points with 2D semantic information, 3D semantic information and fusion semantic information, respectively. Especially, the yellow and red dash ellipses show some of the false and missed detection, respectively. The baseline detector used here is CenterPoint.} 
	\label{Fig:detection result}

\end{figure*}

\subsection{Qualitative Results}

More qualitative detection results are illustrated in Fig. \ref{Fig:detection result}. Fig. \ref{Fig:detection result} (a) shows the annotated ground truth, (b) is the detection results for CenterPoint based on raw point cloud without using any painting strategy. (c) and (d) show the detection results with 2D painted and 3D painted point clouds, respectively, while (e) is the results based on our proposed framework. 
As shown in the figure, there are false positive detection results caused 2D painting due to frustum blurring effect, while 3D painting method produces worse foreground class segmentation compared to  2D image segmentation.  Instead, our FusionPainting can combine two complementary information from 2D and 3D segmentation, and detect objects more accurately.


\subsection{Ablation Studies} \label{secsub:ablation_study}
To verify the effectiveness of different modules, a series of ablation studies have been designed here. All the experiments are executed on the validation split and the settings keep the same as in Sec. \ref{subsec:dataset}. All the results are given in Tab. \ref{tab:ablation_table}, from where we could figure out the impact of each module. 


As demonstrated in Tab. \ref{tab:ablation_table}, 3D semantic segmentation information alone contributes about 9.81\% mAP and 3.92\% NDS averagely, while 2D semantic segmentation information contributes about 21.90\% and 8.46\% averagely. The higher improvement from 2D semantic information benefit from the better recall ability especially for objects in the distance where the point cloud is very sparse. The advantage for 3D semantic information is that the point clouds can well handle the occlusion among objects which is very common and hard to deal in 2D. But the most improvement comes from the combination of 3D Painting, 2D Painting and Adaptive Attention Module, 26.58\% mAP and 10.90\% averagely from 3 detectors we used.



\begin{table}[ht!]
\centering
\resizebox{0.5\textwidth}{!}{%
\begin{tabular}{l | c c c l l }
\hline
\textbf{Baselines} & \textbf{3D-P} & \textbf{2D-P} & \textbf{Att-Mod} & \textbf{mAP(\%)} & \textbf{NDS(\%)} \\ \hline
\multirow{4}{*}{\textbf{Second}}      &  &  &  &50.85 (baseline) &61.96 (baseline)  \\
                 &\checkmark  &  &  &54.45 (\textcolor{red}{+3.60})  &63.41 (\textcolor{red}{+1.45})  \\
                  &  & \checkmark &  &59.40 (\textcolor{red}{+8.55}) &66.03 (\textcolor{red}{+4.07})  \\
                  & \checkmark &  \checkmark &  \checkmark &62.15 (\textcolor{red}{+11.30})  & 67.41 (\textcolor{red}{+5.45})  \\ \hline
\multirow{4}{*}{\textbf{PointPilars}} &  &  &  &43.46 (baseline) &57.50 (baseline)   \\
                      & \checkmark &  &  &52.18 (\textcolor{red}{+8.72})  &61.59 (\textcolor{red}{+4.09})   \\
                     &  & \checkmark  &  &59.10 (\textcolor{red}{+15.64}) & 65.05 (\textcolor{red}{+7.55})  \\
                    & \checkmark &  \checkmark &  \checkmark & 60.77 (\textcolor{red}{+17.31}) &66.04 (\textcolor{red}{+8.54})   \\ \hline
\multirow{4}{*}{\textbf{CenterPoint}} &  &  &  &56.53 (baseline) & 64.82 (baseline) \\
        & \checkmark &  &  &57.83 (\textcolor{red}{+1.30})  &66.32 (\textcolor{red}{+1.50})   \\
            &  & \checkmark &  & 63.82 (\textcolor{red}{+7.29}) & 68.51 (\textcolor{red}{+3.69})\\
         & \checkmark & \checkmark & \checkmark &66.53 (\textcolor{red}{+10.00})  & 70.68 (\textcolor{red}{+5.86}) \\ \hline
\end{tabular}

}
\caption{An ablation study for different fusion strategies. ``Att-Mod'' is short for Adaptive Attention Module, ``3D-P'' and ``2D-P'' are short for ``3D Painting Module'' and ``2D Painting Module'', respectively.}
\label{tab:ablation_table}
\end{table}



\section{Conclusion and Future Works}

We present a novel multi-modal fusion framework, ``FusionPainting'' which can aggregate rich semantic information from 2D/3D semantic segmentation networks. More important, we are the first to use 3D semantic segmentation results as additional information for improving the 3D object detection performance. To fuse both the 2D/3D segmentation results, an adaptive attention module has been proposed to learn an attention mask for each of them. Experimental results on nuScenes dataset illustrate the effectiveness of the proposed strategy. Furthermore, the proposed ``FusionPainting'' is detector independent, which can be freely used for other 3D object detectors. In the future, we plan to deeply integrate the 2D/3D semantic segmentation branches into the detection framework and simultaneously achieve instance segmentation and 3D object detection tasks.

\bibliographystyle{IEEEtran}
\bibliography{IEEEabrv,ref}

\begin{thebibliography}{10}
\providecommand{\url}[1]{#1}
\csname url@samestyle\endcsname
\providecommand{\newblock}{\relax}
\providecommand{\bibinfo}[2]{#2}
\providecommand{\BIBentrySTDinterwordspacing}{\spaceskip=0pt\relax}
\providecommand{\BIBentryALTinterwordstretchfactor}{4}
\providecommand{\BIBentryALTinterwordspacing}{\spaceskip=\fontdimen2\font plus
\BIBentryALTinterwordstretchfactor\fontdimen3\font minus
  \fontdimen4\font\relax}
\providecommand{\BIBforeignlanguage}[2]{{%
\expandafter\ifx\csname l@#1\endcsname\relax
\typeout{** WARNING: IEEEtran.bst: No hyphenation pattern has been}%
\typeout{** loaded for the language `#1'. Using the pattern for}%
\typeout{** the default language instead.}%
\else
\language=\csname l@#1\endcsname
\fi
#2}}
\providecommand{\BIBdecl}{\relax}
\BIBdecl

\bibitem{shi2019pointrcnn}
S.~Shi, X.~Wang, and H.~Li, ``Pointrcnn: 3d object proposal generation and
  detection from point cloud,'' in \emph{Proceedings of the IEEE/CVF Conference
  on Computer Vision and Pattern Recognition}, 2019, pp. 770--779.

\bibitem{zhou2018voxelnet}
Y.~Zhou and O.~Tuzel, ``Voxelnet: End-to-end learning for point cloud based 3d
  object detection,'' in \emph{Proceedings of the IEEE Conference on Computer
  Vision and Pattern Recognition}, 2018, pp. 4490--4499.

\bibitem{lang2019pointpillars}
A.~H. Lang, S.~Vora, H.~Caesar, L.~Zhou, J.~Yang, and O.~Beijbom,
  ``Pointpillars: Fast encoders for object detection from point clouds,'' in
  \emph{Proceedings of the IEEE Conference on Computer Vision and Pattern
  Recognition}, 2019, pp. 12\,697--12\,705.

\bibitem{dou2019seg}
J.~Dou, J.~Xue, and J.~Fang, ``Seg-voxelnet for 3d vehicle detection from rgb
  and lidar data,'' in \emph{2019 International Conference on Robotics and
  Automation (ICRA)}.\hskip 1em plus 0.5em minus 0.4em\relax IEEE, 2019, pp.
  4362--4368.

\bibitem{vora2020pointpainting}
S.~Vora, A.~H. Lang, B.~Helou, and O.~Beijbom, ``Pointpainting: Sequential
  fusion for 3d object detection,'' in \emph{Proceedings of the IEEE/CVF
  Conference on Computer Vision and Pattern Recognition}, 2020, pp. 4604--4612.

\bibitem{zhang2020maff}
Z.~Zhang, M.~Zhang, Z.~Liang, X.~Zhao, M.~Yang, W.~Tan, and S.~Pu, ``Maff-net:
  Filter false positive for 3d vehicle detection with multi-modal adaptive
  feature fusion,'' \emph{arXiv preprint arXiv:2009.10945}, 2020.

\bibitem{xu2018pointfusion}
D.~Xu, D.~Anguelov, and A.~Jain, ``Pointfusion: Deep sensor fusion for 3d
  bounding box estimation,'' in \emph{Proceedings of the IEEE Conference on
  Computer Vision and Pattern Recognition}, 2018, pp. 244--253.

\bibitem{tian2020adaptive}
Y.~Tian, K.~Wang, Y.~Wang, Y.~Tian, Z.~Wang, and F.-Y. Wang, ``Adaptive and
  azimuth-aware fusion network of multimodal local features for 3d object
  detection,'' \emph{Neurocomputing}, vol. 411, pp. 32--44, 2020.

\bibitem{pang2020clocs}
S.~Pang, D.~Morris, and H.~Radha, ``Clocs: Camera-lidar object candidates
  fusion for 3d object detection,'' in \emph{2020 IEEE/RSJ International
  Conference on Intelligent Robots and Systems (IROS)}.\hskip 1em plus 0.5em
  minus 0.4em\relax IEEE, 2020.

\bibitem{ku2018joint}
J.~Ku, M.~Mozifian, J.~Lee, A.~Harakeh, and S.~L. Waslander, ``Joint 3d
  proposal generation and object detection from view aggregation,'' in
  \emph{2018 IEEE/RSJ International Conference on Intelligent Robots and
  Systems (IROS)}.\hskip 1em plus 0.5em minus 0.4em\relax IEEE, 2018, pp. 1--8.

\bibitem{luo2018fast}
W.~Luo, B.~Yang, and R.~Urtasun, ``Fast and furious: Real time end-to-end 3d
  detection, tracking and motion forecasting with a single convolutional net,''
  in \emph{Proceedings of the IEEE conference on Computer Vision and Pattern
  Recognition}, 2018, pp. 3569--3577.

\bibitem{yang2018pixor}
B.~Yang, W.~Luo, and R.~Urtasun, ``Pixor: Real-time 3d object detection from
  point clouds,'' in \emph{Proceedings of the IEEE conference on Computer
  Vision and Pattern Recognition}, 2018, pp. 7652--7660.

\bibitem{liang2020rangercnn}
Z.~Liang, M.~Zhang, Z.~Zhang, X.~Zhao, and S.~Pu, ``Rangercnn: Towards fast and
  accurate 3d object detection with range image representation,'' \emph{arXiv
  preprint arXiv:2009.00206}, 2020.

\bibitem{hu2020you}
P.~Hu, J.~Ziglar, D.~Held, and D.~Ramanan, ``What you see is what you get:
  Exploiting visibility for 3d object detection,'' in \emph{Proceedings of the
  IEEE/CVF Conference on Computer Vision and Pattern Recognition}, 2020, pp.
  11\,001--11\,009.

\bibitem{yan2018second}
Y.~Yan, Y.~Mao, and B.~Li, ``Second: Sparsely embedded convolutional
  detection,'' \emph{Sensors}, vol.~18, no.~10, p. 3337, 2018.

\bibitem{kuang2020voxel}
H.~Kuang, B.~Wang, J.~An, M.~Zhang, and Z.~Zhang, ``Voxel-fpn: Multi-scale
  voxel feature aggregation for 3d object detection from lidar point clouds,''
  \emph{Sensors}, vol.~20, no.~3, p. 704, 2020.

\bibitem{yin2020lidar}
J.~Yin, J.~Shen, C.~Guan, D.~Zhou, and R.~Yang, ``Lidar-based online 3d video
  object detection with graph-based message passing and spatiotemporal
  transformer attention,'' in \emph{Proceedings of the IEEE/CVF Conference on
  Computer Vision and Pattern Recognition}, 2020, pp. 11\,495--11\,504.

\bibitem{zhou2019iou}
D.~Zhou, J.~Fang, X.~Song, C.~Guan, J.~Yin, Y.~Dai, and R.~Yang, ``Iou loss for
  2d/3d object detection,'' in \emph{2019 International Conference on 3D Vision
  (3DV)}.\hskip 1em plus 0.5em minus 0.4em\relax IEEE, 2019, pp. 85--94.

\bibitem{zhou2020joint}
D.~Zhou, J.~Fang, X.~Song, L.~Liu, J.~Yin, Y.~Dai, H.~Li, and R.~Yang, ``Joint
  3d instance segmentation and object detection for autonomous driving,'' in
  \emph{Proceedings of the IEEE/CVF Conference on Computer Vision and Pattern
  Recognition}, 2020, pp. 1839--1849.

\bibitem{qi2018frustum}
C.~R. Qi, W.~Liu, C.~Wu, H.~Su, and L.~J. Guibas, ``Frustum pointnets for 3d
  object detection from rgb-d data,'' in \emph{Proceedings of the IEEE
  Conference on Computer Vision and Pattern Recognition}, 2018, pp. 918--927.

\bibitem{yin2021center}
T.~Yin, X.~Zhou, and P.~Kr{\"a}henb{\"u}hl, ``Center-based 3d object detection
  and tracking,'' \emph{Proceedings of the IEEE Conference on Computer Vision
  and Pattern Recognition}, 2021.

\bibitem{duan2019centernet}
K.~Duan, S.~Bai, L.~Xie, H.~Qi, Q.~Huang, and Q.~Tian, ``Centernet: Keypoint
  triplets for object detection,'' in \emph{Proceedings of the IEEE/CVF
  International Conference on Computer Vision}, 2019, pp. 6569--6578.

\bibitem{graham20183d}
B.~Graham, M.~Engelcke, and L.~Van Der~Maaten, ``3d semantic segmentation with
  submanifold sparse convolutional networks,'' in \emph{Proceedings of the IEEE
  conference on computer vision and pattern recognition}, 2018, pp. 9224--9232.

\bibitem{qi2017pointnet}
C.~R. Qi, H.~Su, K.~Mo, and L.~J. Guibas, ``Pointnet: Deep learning on point
  sets for 3d classification and segmentation,'' in \emph{Proceedings of the
  IEEE Conference on Computer Vision and Pattern Recognition}, 2017, pp.
  652--660.

\bibitem{shi2020pv}
S.~Shi, C.~Guo, L.~Jiang, Z.~Wang, J.~Shi, X.~Wang, and H.~Li, ``Pv-rcnn:
  Point-voxel feature set abstraction for 3d object detection,'' in
  \emph{Proceedings of the IEEE/CVF Conference on Computer Vision and Pattern
  Recognition}, 2020, pp. 10\,529--10\,538.

\bibitem{chen2017multi}
X.~Chen, H.~Ma, J.~Wan, B.~Li, and T.~Xia, ``Multi-view 3d object detection
  network for autonomous driving,'' in \emph{Proceedings of the IEEE Conference
  on Computer Vision and Pattern Recognition}, 2017, pp. 1907--1915.

\bibitem{he2017mask}
K.~He, G.~Gkioxari, P.~Doll{\'a}r, and R.~Girshick, ``Mask r-cnn,'' in
  \emph{Proceedings of the IEEE international conference on computer vision},
  2017, pp. 2961--2969.

\bibitem{liang2019multi}
M.~Liang, B.~Yang, Y.~Chen, R.~Hu, and R.~Urtasun, ``Multi-task multi-sensor
  fusion for 3d object detection,'' in \emph{Proceedings of the IEEE/CVF
  Conference on Computer Vision and Pattern Recognition}, 2019, pp. 7345--7353.

\bibitem{yang2018hdnet}
B.~Yang, M.~Liang, and R.~Urtasun, ``Hdnet: Exploiting hd maps for 3d object
  detection,'' in \emph{Conference on Robot Learning}.\hskip 1em plus 0.5em
  minus 0.4em\relax PMLR, 2018, pp. 146--155.

\bibitem{fang2021mapfusion}
J.~Fang, D.~Zhou, X.~Song, and L.~Zhang, ``Mapfusion: A general framework for
  3d object detection with hdmaps,'' \emph{arXiv preprint arXiv:2103.05929},
  2021.

\bibitem{chen2018encoder}
L.-C. Chen, Y.~Zhu, G.~Papandreou, F.~Schroff, and H.~Adam, ``Encoder-decoder
  with atrous separable convolution for semantic image segmentation,'' in
  \emph{Proceedings of the European conference on computer vision (ECCV)},
  2018, pp. 801--818.

\bibitem{zhao2017pyramid}
H.~Zhao, J.~Shi, X.~Qi, X.~Wang, and J.~Jia, ``Pyramid scene parsing network,''
  in \emph{Proceedings of the IEEE conference on computer vision and pattern
  recognition}, 2017, pp. 2881--2890.

\bibitem{chen2019hybrid}
K.~Chen, J.~Pang, J.~Wang, Y.~Xiong, X.~Li, S.~Sun, W.~Feng, Z.~Liu, J.~Shi,
  W.~Ouyang \emph{et~al.}, ``Hybrid task cascade for instance segmentation,''
  in \emph{Proceedings of the IEEE/CVF Conference on Computer Vision and
  Pattern Recognition}, 2019, pp. 4974--4983.

\bibitem{shelhamer2017fully}
E.~Shelhamer, J.~Long, and T.~Darrell, ``Fully convolutional networks for
  semantic segmentation,'' \emph{IEEE transactions on pattern analysis and
  machine intelligence}, vol.~39, no.~4, pp. 640--651, 2017.

\bibitem{Howard_2019_ICCV}
A.~Howard, M.~Sandler, G.~Chu, L.-C. Chen, B.~Chen, M.~Tan, W.~Wang, Y.~Zhu,
  R.~Pang, V.~Vasudevan, Q.~V. Le, and H.~Adam, ``Searching for mobilenetv3,''
  in \emph{The IEEE International Conference on Computer Vision (ICCV)},
  October 2019, pp. 1314--1324.

\bibitem{cong2021input}
P.~Cong, X.~Zhu, and Y.~Ma, ``Input-output balanced framework for long-tailed
  lidar semantic segmentation,'' \emph{arXiv preprint arXiv:2103.14269}, 2021.

\bibitem{zhang2020polarnet}
Y.~Zhang, Z.~Zhou, P.~David, X.~Yue, Z.~Xi, B.~Gong, and H.~Foroosh,
  ``Polarnet: An improved grid representation for online lidar point clouds
  semantic segmentation,'' in \emph{Proceedings of the IEEE/CVF Conference on
  Computer Vision and Pattern Recognition}, 2020, pp. 9601--9610.

\bibitem{cheng20212}
R.~Cheng, R.~Razani, E.~Taghavi, E.~Li, and B.~Liu, ``2-s3net: Attentive
  feature fusion with adaptive feature selection for sparse semantic
  segmentation network,'' \emph{arXiv preprint arXiv:2102.04530}, 2021.

\bibitem{xu2021rpvnet}
J.~Xu, R.~Zhang, J.~Dou, Y.~Zhu, J.~Sun, and S.~Pu, ``Rpvnet: A deep and
  efficient range-point-voxel fusion network for lidar point cloud
  segmentation,'' \emph{arXiv preprint arXiv:2103.12978}, 2021.

\bibitem{yang20203dssd}
Z.~Yang, Y.~Sun, S.~Liu, and J.~Jia, ``3dssd: Point-based 3d single stage
  object detector,'' in \emph{Proceedings of the IEEE/CVF conference on
  computer vision and pattern recognition}, 2020, pp. 11\,040--11\,048.

\bibitem{zhu2019class}
B.~Zhu, Z.~Jiang, X.~Zhou, Z.~Li, and G.~Yu, ``Class-balanced grouping and
  sampling for point cloud 3d object detection,'' \emph{arXiv preprint
  arXiv:1908.09492}, 2019.

\bibitem{chen2020object}
Q.~Chen, L.~Sun, Z.~Wang, K.~Jia, and A.~Yuille, ``Object as hotspots: An
  anchor-free 3d object detection approach via firing of hotspots,'' in
  \emph{European Conference on Computer Vision}.\hskip 1em plus 0.5em minus
  0.4em\relax Springer, 2020, pp. 68--84.

\bibitem{chen2020every}
Q.~Chen, L.~Sun, E.~Cheung, and A.~L. Yuille, ``Every view counts: Cross-view
  consistency in 3d object detection with hybrid-cylindrical-spherical
  voxelization,'' \emph{Advances in Neural Information Processing Systems},
  vol.~33, 2020.

\bibitem{DBLP:conf/eccv/YooKKC20}
\BIBentryALTinterwordspacing
J.~H. Yoo, Y.~Kim, J.~S. Kim, and J.~W. Choi, ``3d-cvf: Generating joint camera
  and lidar features using cross-view spatial feature fusion for 3d object
  detection,'' in \emph{Computer Vision - {ECCV} 2020 - 16th European
  Conference, Glasgow, UK, August 23-28, 2020, Proceedings, Part {XXVII}}, ser.
  Lecture Notes in Computer Science, A.~Vedaldi, H.~Bischof, T.~Brox, and
  J.~Frahm, Eds., vol. 12372.\hskip 1em plus 0.5em minus 0.4em\relax Springer,
  2020, pp. 720--736. [Online]. Available:
  \url{https://doi.org/10.1007/978-3-030-58583-9\_43}
\BIBentrySTDinterwordspacing

\bibitem{nuscenes2019}
H.~Caesar, V.~Bankiti, A.~H. Lang, S.~Vora, V.~E. Liong, Q.~Xu, A.~Krishnan,
  Y.~Pan, G.~Baldan, and O.~Beijbom, ``nuscenes: A multimodal dataset for
  autonomous driving,'' \emph{arXiv preprint arXiv:1903.11027}, 2019.

\bibitem{everingham2010pascal}
M.~Everingham, L.~Van~Gool, C.~K. Williams, J.~Winn, and A.~Zisserman, ``The
  pascal visual object classes (voc) challenge,'' \emph{International journal
  of computer vision}, vol.~88, no.~2, pp. 303--338, 2010.

\bibitem{zhou2020cylinder3d}
H.~Zhou, X.~Zhu, X.~Song, Y.~Ma, Z.~Wang, H.~Li, and D.~Lin, ``Cylinder3d: An
  effective 3d framework for driving-scene lidar semantic segmentation,''
  \emph{arXiv preprint arXiv:2008.01550}, 2020.

\bibitem{loshchilov2019decoupled}
\BIBentryALTinterwordspacing
I.~Loshchilov and F.~Hutter, ``Decoupled weight decay regularization,'' in
  \emph{International Conference on Learning Representations}, 2019. [Online].
  Available: \url{https://openreview.net/forum?id=Bkg6RiCqY7}
\BIBentrySTDinterwordspacing

\end{thebibliography}
\end{document}